\documentclass[journal]{IEEEtran}
\usepackage{multicol}
\usepackage{subfig}
\usepackage{stfloats}
\usepackage{hyperref}
\usepackage{url}

\usepackage {graphicx}
\usepackage{array}
\usepackage{amssymb}
\usepackage{dsfont}
\usepackage{cite}
\usepackage[square, comma, sort&compress, numbers]{natbib}

\ifCLASSINFOpdf
\else
\fi
\begin{document}

%
\title{Perceptual uniform descriptor and Ranking on manifold: A bridge between
image representation and ranking for image retrieval}
%
%
%

\author{Shenglan~Liu,
        Jun~Wu,
        Lin~Feng,
        Yang~Liu,
        ~Hong~Qiao,~\IEEEmembership{Senior~Member,~IEEE}
        , Wenbo Luo Muxin Sun, and ~Wei~Wang,~\IEEEmembership{Senior~Member,~IEEE}
\thanks{Shenglan Liu,
Lin Feng, Yang Liu and Wei Wang are with Faculty of Electronic Information and Electrical Engineering, Dalian University of Technology, Dalian, Liaoning, 116024 China. Jun Wu is with the School of Innovation and Entrepreneurship, Dalian University of Technology, Dalian, Liaoning, 116024 China. e-mail: (\{liusl,dlut\_liuyang,shuxuewujun\}@mail.dlut.edu.cn, \{fenglin,wangwei\}@dlut.edu.cn). Lin Feng is the Corresponding author.}
\thanks{Wenbo Luo is with the School of Psychology, Liaoning Normal University, Dalian, China e-mail: (wenbo9390@sina.com)}
\thanks{This work was supported by National Natural Science Foundation of
P.R. China (61602082, 61370200, 61672130).}}
\markboth{Journal of \LaTeX\ Class Files,~Vol.~, No.~,}%
{Shell \MakeLowercase{\textit{et al.}}: Bare Demo of IEEEtran.cls for Journals}
%



\maketitle

\begin{abstract}
Incompatibility of image descriptor and ranking is always neglected in image
retrieval. In this paper, manifold learning and Gestalt psychology theory
are involved to solve the incompatibility problem. A new holistic
descriptor called Perceptual Uniform Descriptor (PUD) based on Gestalt psychology
is proposed, which combines color and gradient direction to imitate the
human visual uniformity. PUD features in the same class images distributes on one
manifold in most cases because PUD improves the visual uniformity of the
traditional descriptors. Thus, we use manifold ranking and PUD to realize
image retrieval. Experiments were carried out on five benchmark data sets,
and the proposed method can greatly improve the accuracy of image
retrieval. Our experimental results in the Ukbench and Corel-1K datasets demonstrated that N-S score reached to 3.58 (HSV 3.4) and mAP to 81.77\% (ODBTC 77.9\%) respectively by utilizing PUD which has only 280 dimension. The results are higher than other holistic image
descriptors (even some local ones) and state-of-the-arts retrieval methods.
\end{abstract}


%
\IEEEpeerreviewmaketitle

\section{Introduction}
%
%
%
%
\IEEEPARstart{F}{eature} extraction and ranking are two important topics in content based
image retrieval (CBIR). A significant number of excellent methods for image
feature extraction and ranking have been proposed in recent years \cite{1,2}.

It is well known that image representation plays an important part in CBIR
systems \cite{4,5}, and thus the performance of these systems depends mainly on
the discrimination and effectiveness of features. Many research works have
already provided excellent image descriptors according to the different
understanding for image data. The process of feature extracted can be
divided into three steps: 1) image preprocessing; 2) the detection of
discriminative image regions; 3) feature statistical strategy in these
regions. Most of the works concentrated on one or more steps to improve
their descriptors.

First, in order to describe certain properties of natural
images which may contain various types of image noise, image preprocessing
is an indispensable step. Many image denoising \cite{6} and image sharpening
algorithms \cite{7} have been presented to reduce the effect of noise on image
content and strengthen discriminative information in some regions. In
addition, color is transformed to gray in natural images \cite{8}, and then
texture and sharp can be described regardless of the disturbance of color.

Second, discriminative image regions are detected. Based on that, the
descriptors can be classified into global-based and local-based. Color
Histogram (CH) \cite{9}, Local Binary Patterns (LBP) \cite{10,11} and Histogram of
Gradient (HOG) \cite{12}, which describe the color, texture and edge features
respectively, are provided based on the global image regions. Motivated by
the visual perception mechanisms for image retrieval, Liu et al. provided
micro-structure descriptor (MSD) \cite{13} which defined the micro-structures
through the similarity of edge orientation and the underlying colors, and
introduced structure element correlation statistics to characterize the
spatial correlation among them. And color difference histogram (CDH)
\cite{14} characterized visual perceptual differences by uniform color difference
in the global image, which also demonstrated superior performance in CBIR.
On the contrary, local-based descriptors focus on describing local regions
which contain certain information. Lowe et al. \cite{15} introduced a local
descriptor called scale-invariant feature transform (SIFT), which aimed at
detecting and describing some local neighborhoods around key points in scale
space. Due to the similarity with the receptive field of the mammalian
cortical simple cells, Gabor wavelets \cite{16} have been applied to image
analysis. HMAX model \cite{17} based on the hierarchical visual processing in
the primary visual cortex (V1) utilized Gabor filters in different
scales and orientations in S1 unit. More details about performance comparisons among other local descriptors are presented in \cite{18}.

Finally, corresponding feature statistics methods in these regions are provided. As
one of the most common methods, histogram-based strategy has been
applied in many descriptors, such as CH, LBP and HOG.
This strategy emphasizes the occurrence probability of certain element and
thus it is effective and easy to realize. However, considering the loss of
spatial relationship among these elements, the performance of this strategy
is limited. To solve the problems in CH, color moment \cite{19}, color
correlogram \cite{20} and color coherence vector \cite{21} were proposed to obtain both
the holistic distribution and spatial correlation among pixels.

Besides image feature extraction methods, image Ranking has also been rapidly
developed in CBIR. Lots of researches have been devoted in improving the
ranking results, such as $L1$-norm \cite{22}, Euclidean distance \cite{23}, Harmming
distance \cite{24} etc. Previous researches showed that ranking by $L1$-norm is simple and can obtain a better result than that by Euclidean distance \cite{25,26}. In
addition, the graph based ranking methods, such as PageRank \cite{27} and
manifold ranking \cite{28}, are also widely used for image retrieval. PageRank
and manifold ranking without queries can yield the same ranking list.
Manifold ranking is proposed based on manifold learning and relates to
perception.

In most image retrieval schemes, image feature extraction and ranking are
two independent processes \cite{29}. This likely accounts for the incompatibility between descriptor and ranking method (for example, an image representation which is compatible with 1-norm ranking, may not obtain expectant results while using manifold ranking
methods, see section  \uppercase\expandafter{\romannumeral 7}).

In computer vision, we hope the computer to imitate human's perception for
learning image and other visual data \cite{30}. In the process of human
cognition, visual uniformity is beneficial to learn image, and has been
used for the extraction of the image features \cite{31}. Visual uniformity is
consistent with human perception of image. Thus, we point out that the image
features extraction by visual uniformity are more likely to distribute on
the manifold. In 2000, three researches related to manifold learning were published in ``Science'' \cite{32,33,34}, in which Lee \cite{32} points out that ``human perception
is in the way of manifold'' (This phenomenon is illustrated in section  \uppercase\expandafter{\romannumeral 2}). In
this paper, we construct the image feature and ranking model based on
manifold, which aims to realize the uniformity in CBIR. Our method improves the
efficiency and accuracy of retrieval results, and avoids the incompatible
problem of the descriptor and ranking in image retrieval.

In this paper, according to visual organization principle and the theory in
``The manifold way of perception'', we use human's visual perception to
construct the image visual feature, and retrieve images via manifold ranking. The main contributions of this paper are stated as follows:

(1) Perceptual Uniform Descriptor (PUD) is proposed by using the visual principle of
Gestalt psychology, so that it can better distribute on a manifold.

(2) The incompatible problem between image descriptors and ranking methods is
analyzed. The concept of manifold is involved as a bridge for descriptors
and ranking methods in CBIR.

The rest of the paper is organized as follows: Section  \uppercase\expandafter{\romannumeral 2} states the
motivation of our proposed image retrieval scheme. Principles of Gestalt
psychology are introduced in section \uppercase\expandafter{\romannumeral 3}. Section  \uppercase\expandafter{\romannumeral 4} and section  \uppercase\expandafter{\romannumeral 5} present our image descriptor. Section  \uppercase\expandafter{\romannumeral 6} refers to manifold ranking for image retrieval.
In section  \uppercase\expandafter{\romannumeral 7}, experimental results and analysis are reported. Section  \uppercase\expandafter{\romannumeral 8}
concludes the paper.

\section{Motivation}
Human visual system can pinpoint and analyze objects in complex images in a
very short time. The main aims of many studies related to human brain visual
mechanism and cognitive psychology are to simulate vision systems that have
the equal performance to humans in object recognition. According to the
analysis that the image variability can actually be considered as a manifold
embedded in the image space, Seung and Lee introduced the idea that human
visual perception can be expressed by manifolds. The brain must encode the
visual information by some ways. For image analysis, the descriptors
that are in accordance with the distribution of manifolds have more
discriminative information.

Due to the connection with low-level visual features, human
visual attention system related to the perception and
understanding for visual images facilitates the construction of image feature
representation. Some psychophysical and neurobiological studies demonstrate
that human visual system is sensitive to the low-level visual features, such
as color and edge information \cite{35}. However, the holistic images usually
contain some redundant regions where less discriminative information is
useful for image analysis. And the image representation in these regions may
not only impair the performance of descriptor but also consume too much
time. To detect the special regions that human eyes perceive predominantly,
the studies in the cognitive psychology give some inspiration about
perceiving the objects. The Gestalt Laws of perception introduces some
principles that help people group similar pixels or patches in image. Among these
principles, proximity, similarity and good continuation are fundamental to
define the perceptually uniform regions. The closer the image pixels are
or the more similar their low-level features are, the more likely it is that
they belong to the same region according to the law of proximity and similarity. And the good
continuation among pixels can characterize the contour and edge in these
regions.

Motivated by the manifold ways of perception and the Gestalt Laws of
perception for visual images, this paper presents a novel image descriptor
called Perceptual Uniform Descriptor (PUD), which characterizes the
discriminative information in the visual perceptually uniform regions. Based
on the three principles in the Gestalt Laws of perception, perceptually
uniform regions are defined as the local regions where neighboring pixels have similar low-level visual features. Then image discriminative information can
be characterized by two orthogonal properties: spatial structure and
contrast in these regions. Considering that both local and holistic distributions among the pixels are
significant for describing image content, we propose the color difference feature which fuses
color difference correlation and global color difference histogram, and
texton frequency feature which fuses texton frequency correlation and texton
frequency histogram to represent contrast and spatial structure information,
respectively.

\section{Principles of Gestalt psychology}

Gestalt psychology \cite{37}, which is designed based on the understanding for human
visual perception, allows visually similar objects to be grouped into unity.
And this idea implies that ``the whole is greater than sum of the parts''.
The principles of Gestalt psychology are highly relevant to the perception for the world, and can be applied to help design visual communication models.
This paper focuses on three main principles in Gestalt psychology, namely proximity,
similarity and good continuation.

\subsection{The Gestalt Law of Proximity}
\begin{figure}[htbp]
\centerline{
\includegraphics[width=3.5in]{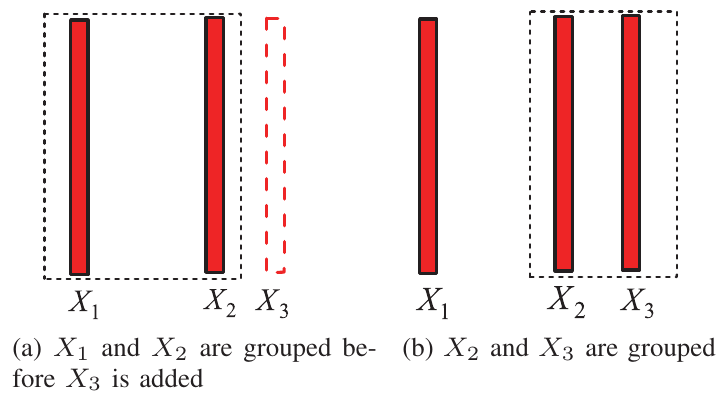}}
\caption{The Gestalt Law of Proximity}
\label{fig2}
\end{figure}

The law of proximity suggests that elements which are close in position are
easy to be grouped. Based on that, the perceptual model is presented as
follows:

\begin{equation}
\label{eq1}
\left[ {X_1 \cdot X_2 } \right] \cdot X_3 \to X_1 \cdot \left[ {X_2 \cdot
X_3 } \right]
\end{equation}
where $X_1$, $X_2$ and $X_3$ denote three separate elements which may be
grouped according to the proximity. As illustrated in Fig. \ref{fig2}, $X_1$
and $X_2$ are grouped at first. After $X_3$ is added, $X_2$ and
$X_3$ are more likely to be grouped due to their proximity. As discussed
above, the characteristic that the closer these elements are the more
likely they are to be perceived as a group, can be applied to visually
communicate through concentrating on key elements.

\subsection{The Gestalt Law of Similarity}

\begin{figure}[htbp]
\centerline{
\includegraphics[width=3.5in]{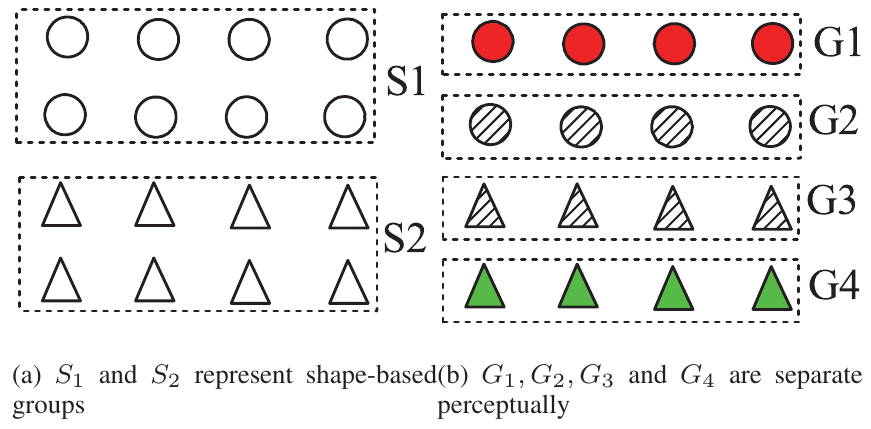}}
\caption{The Gestalt law of similarity}
\label{fig3}
\end{figure}

Elements tend to be perceived as a group when they have similar property,
such as color, texture, shape, etc. This law emphasizes the similarity among
elements, which can be described as:

\begin{equation}
\label{eq2}
\left[ {W_1 \cdot X_1 \cdot Y_1 \cdot Z_1 } \right] \ne \left[ {W_2 \cdot X_2 \cdot Y_2 \cdot
Z_2 } \right] \ne \left[ {W_3 \cdot X_3 \cdot Y_3 \cdot Z_3 } \right]
\end{equation}
where $W$, $X$, $Y$ and $Z$ are different elements which share the same property,
namely color $\left( {_1 } \right)$, texture $\left( {_2 } \right)$ or
shape$\left( {_3 } \right)$. As shown in Fig. \ref{fig3}, both $S_1 $ and $S_2 $ are
groups that have similar shape, but when their color or texture are
different significantly $(( {G_1 ,G_2 }), ({G_3 ,G_4 }))$, elements may be distinguished even though they share similar
shape. In addition, this principle may over-ride the law of proximity. In other
words, similar property and proximity act together as catalysts in
perceiving groups.

\subsection{The Gestalt Law of Good Continuation}
\begin{center}
\begin{figure}[htbp]
\centerline{\includegraphics[width=1.5in]{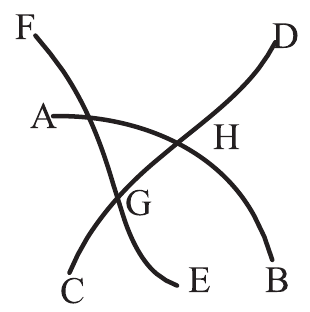}}
\caption{The Gestalt law of good continuation}
\label{fig4}
\end{figure}
\end{center}

This law suggests that elements which share certain lines, curves and planes
are likely to be perceived as a continuous object. Though one element may be
divided into some parts, it is a unity due to the good continuation. In a
sense, the whole element is more favorable to reconstruct the object than
several small parts which are interconnected with each other. This law can be
depicted as follows:
\begin{equation}
\label{eq3}
\left[ {X_1 + X_2 + X_3 } \right] = \left[ X \right]
\end{equation}
where $X_1$, $X_2$ and $X_3$ are three separate visual elements and $X$ is a
new visual unity which is perceptually grouped due to the law of good
continuation. Take Fig. \ref{fig4} as an example, $X_1$ , $X_2$ and $X_3$ denote three
separate lines CG, GH and HD, respectively. It can be seen that these lines are likely to be
perceptually grouped as a continual line CD though they seem to be cut off
by the lines AB and EF. Thus it demonstrates that line CD is more likely to
character the object than three separate continual parts for human visual
system.

\section{Perceptually uniform regions}
In this paper, perceptually uniform regions are defined as local image
regions where pixels have similar property with their neighbors.
According to the law of proximity and similarity in the Gestalt Laws of
perception, pixels that are close to each other or have the similar
property are more likely to be grouped into unity. The comparison and
detection can be processed in the patterns with fixed size. As Julesz's textons
theory claimed \cite{38,39}, image can be seen as the formation of regular
structural elements. Though natural images are difficult to meet the
requirement of regular structural elements, the idea that local regular
patterns are used as the medium to discern certain regions can still be
adopted. In this section, perceptually uniform regions can be detected by
certain small blocks. In a block with a grid of size $3\times3$, the central pixel
can be compared with its all neighbors to identify whether this block
belongs to perceptually uniform region. The proximity between the central
pixels and its neighbors has implied that they are likely
to belong to the same region. And the similarity in property between them
further proves that conclusion. Besides that, good continuation not only represents the
integrity of the region, but also characterizes the contour and edge.

Due to the sensitivity of visual system to color and edge orientation, the
two properties are extracted to detect the perceptually uniform regions respectively.
Even though there are many excellent color space models \cite{40}, HSV color
space is chosen to describe the color attribute because it is close to the
visual perception of human eyes. An image may have thousands kinds of color
values, thus it is time-consuming to measure the similarity of pixels. The
HSV color space is uniformly quantized into 128 bins, and H, S and V
channels are divided into 8, 4 and 4 bins respectively. To describe the edge
orientation of color images, Di Zenzo \cite{41} had proposed a method to obtain
gradient value and orientation using color image directly, which avoided the
loss of color information in traditional methods utilizing gray images. To
express the main difference, gradient orientation is also uniformly
quantized into 12 bins. Next, the color and gradient orientation of pixels
are compared to identify the perceptually uniform regions.

\begin{center}
\begin{figure}[htbp]
\centerline{\includegraphics[width=3.8in]{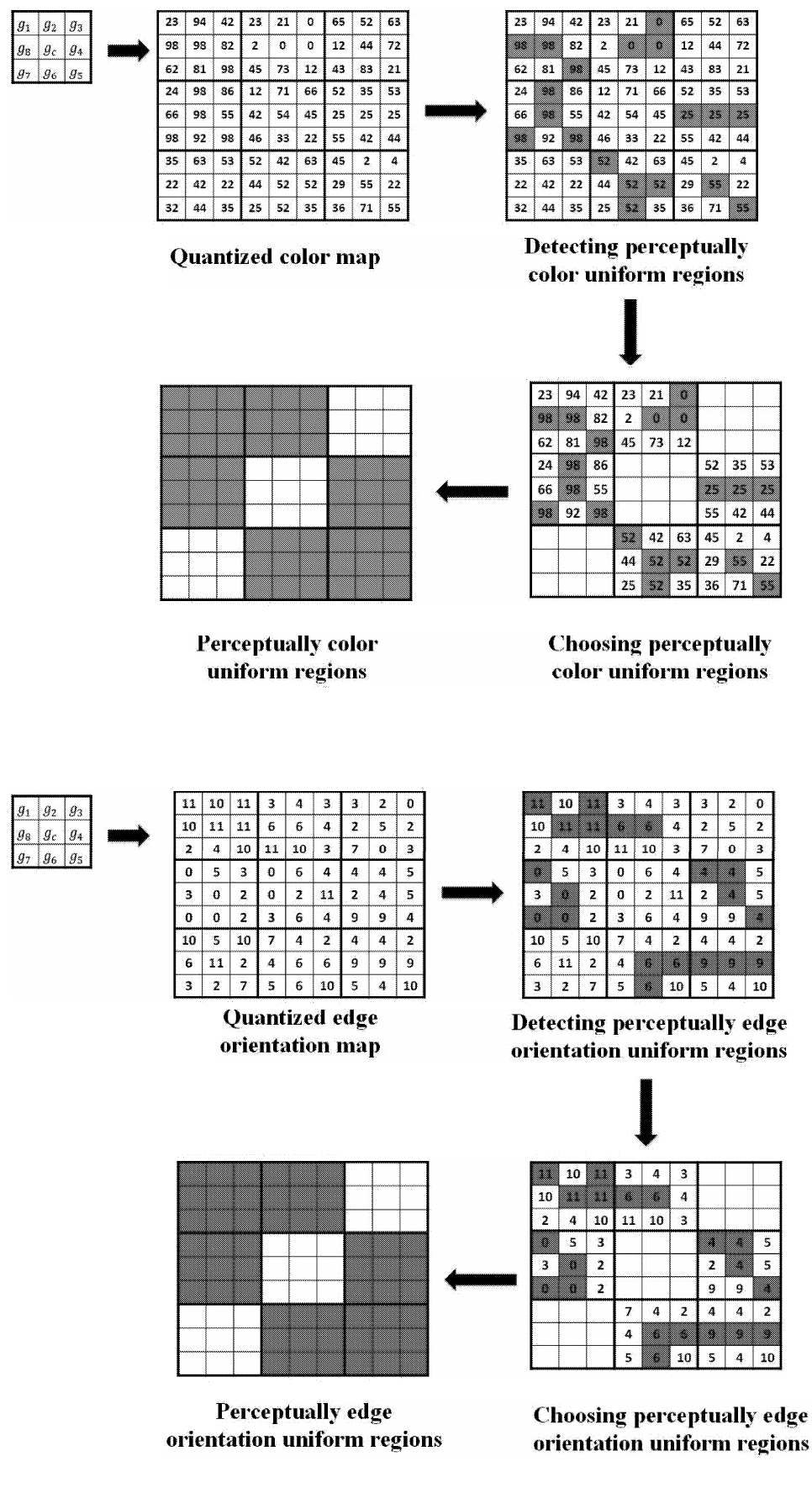}}
\caption{The detecting process for perceptually uniform regions (the green one is color uniform regions detection, and the red one means edge orientation uniform regions detection)}
\label{fig5}
\end{figure}
\end{center}

Considering that there are two properties of images, perceptually uniform
regions should be detected in the quantized color and gradient orientation
map, respectively. Take quantized color map for example, through comparing a
pixel with its neighbors, if they have similar quantized color value, it means that they are likely to be grouped perceptually due to
the law of proximity and similarity. Thus if at least one neighboring pixel
have the same property, it demonstrates that this block can be identified as
the perceptually uniform regions. As illustrated in Fig. \ref{fig5}, the detecting
strategy in this section can be described as the following four steps in details:

(1) Convert the original RGB to HSV color space, and then compute its quantized color and
edge orientation map as discussed above;

(2) Detect the similarity between the central pixel and its neighbors in $3\times3$
block, and slide the block to ensure that the surrounding regions of all the
pixels are analyzed except the marginal pixels which have little
discriminative information;

(3) Choose the regions where there are neighbors sharing similar property with center pixel;

(4) The regions chosen in the last step are conserved as the perceptual uniform
regions, and obtain color-based and edge-based uniform regions to
characterize respectively.

\begin{center}
\begin{figure}[htbp]
\centerline{\includegraphics[width=3.5in]{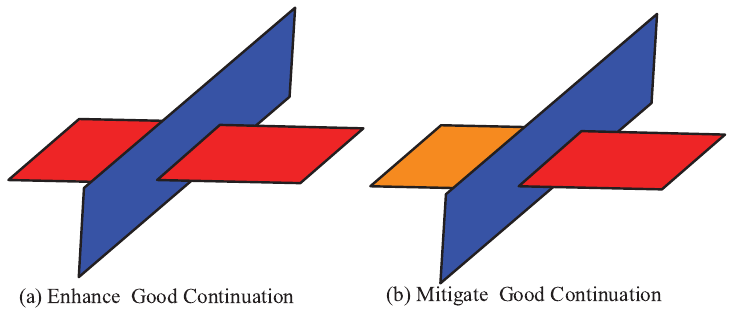}}
\caption{The good continuation based on color property}
\label{fig6}
\end{figure}
\end{center}

Due to the law of good continuation, the pixels that have similar
property can be perceptually grouped as the whole object. It is in favor of
characterizing the holistic feature of the object. For example, the
similar color can enhance the good continuation in Fig. \ref{fig6}(a), and thus the
two red planes are easily perceived as a whole. In contrast, the dissimilar
color mitigates the good continuation in Fig. \ref{fig6}(b). The same conclusion can
be made when considering the edge orientation property. The details of
feature extraction method in these regions is described in next section.

\section{Perceptually uniform feature representation}

Image can be regarded as a collection of pixels. Spatial structure and
contrast are important and orthogonal features where spatial structure is
the correlation among pixels and contrast represents the difference of
pixels. Perceptually uniform color difference have shown superior
performance in CBIR \cite{14}. The Euclidean distance between two pixels in color
space measures the degree of visual perceptual difference. Even though
neighboring pixels may have identical quantized color and gradient
orientation, their slight difference are still important in discriminating
natural images. Besides that, the great difference among other pixels also
provide much power to analyze the contrast of images. A correlation
statistical strategy \cite{13} based on structure element have been applied to
express the spatial structure among neighboring pixels. For local regions,
this method provide certain positional relationship which makes up for
the limitation of contrast information. However, there still exist some
problems to be solved in holistic regions. Based on these, we propose color
difference feature and texton frequency feature to characterize the
perceptually uniform images, respectively.

\subsection{Color difference feature}
Color difference feature aims to describe contrast information in the local
patterns and holistic regions simultaneously. For a color image $f\left( {x,y}
\right)$, its corresponding perceptually uniform images can be denoted as
$T_k \left( {k = 1,2} \right)$ where $T_1 \left( {x,y} \right) \in \left\{
{0,1,...,127} \right\}$ denotes the quantized color map and
$T_2 \left( {x,y} \right) \in \left\{ {0,1,...,11} \right\}$ denotes the quantized edge orientation map. Before measuring the color
difference, it is necessary to preprocess the color space. HSV color space
is introduced based on the cylinder coordinate system and it is unreasonable
to equally measure the three components H, S and V. So this color space need
to be transformed to the Cartesian coordinate system ${H}'{S}'{V}'$ where
${H}' = S \cdot \cos \left( H \right),{S}' = S \cdot \sin \left( H \right)$
and ${V}' = V$. Then through computing the Euclidean distance among pixels in
this new system, perceptually uniform color difference is expressed.
Considering a $3\times3$ block, the color difference between the central pixel
$\mbox{g}_c = \left( {x_c ,y_c } \right)$ and its neighbors $\mbox{g}_i = \left(
{x_i ,y_i } \right)$ is measured as:

\begin{equation}
\label{eq4}
d_i = \left\{ {\begin{array}{l}
 \sqrt {\left( {\Delta {H}'_i } \right)^2 + \left( {\Delta {S}'_i }
\right)^2 + \left( {\Delta {V}'_i } \right)^2} \\
 where ~~~ \max \left( {\left| {x_c - x_i } \right|,\left| {y_c - y_i } \right|}
\right) = D \\
 \end{array}} \right.
\end{equation}

In this paper, we only choose the nearest pixels as neighbors of the center pixel because the time complexity increases greatly when considering more neighbors. So we set $D=1$ to select eight neighbors $\mbox{g}_i \left( {i =0,1,...,7} \right)$ in $3\times3$ block around the center pixel $\mbox{g}_c $. To characterize
local and holistic color difference property, we firstly extract color
difference correlation which aims to describe the contrast in local region,
and global color difference histogram which expresses the global
distribution of contrast. Then a feature fusing method is used to obtain the
advantages of them, and meanwhile avoid the expansion of feature dimension.

Color difference correlation is defined as the color difference distribution
that two neighboring pixels have the similar properties. $D\left( {T_k
\left( {\mbox{g}_c } \right)} \right)$ denotes the sum of color difference when
neighbors $\mbox{g}_i $ have the similar property $T_k $ with $\mbox{g}_c $ and $\bar
{D}\left( {T_k \left( {\mbox{g}_c } \right)} \right)$ denotes the sum of all
possible color difference. Thus, $D$ and $\bar{D}$ can be expressed as follows:
\begin{equation}
\label{eq5}
D\left( {T_k \left( {\mbox{g}_c } \right)} \right) = \sum\nolimits_i {\delta \left(
{\mbox{g}_c ,\mbox{g}_i } \right) \cdot d_i }
\end{equation}

\begin{equation}
\label{eq6}
\bar {D}\left( {T_k \left( {\mbox{g}_c } \right)} \right) = \sum\nolimits_i {d_i }
\end{equation}
where $\delta _c \left( {\mbox{g}_c ,\mbox{g}_i } \right)$ is a discriminant function to identify
whether the neighboring pixels are similar, and can be written as:
\begin{equation}
\label{eq7}
\delta \left( {\mbox{g}_c ,\mbox{g}_i } \right) = \left\{ {\begin{array}{l}
 1,\;\;\;\;\;\;\;T_k \left( {\mbox{g}_c } \right) = T_k \left( {\mbox{g}_i } \right) \\
 0,\;\;\;\;\;\;\;T_k \left( {\mbox{g}_c } \right) \ne T_k \left( {\mbox{g}_i } \right) \\
 \end{array}} \right.
\end{equation}

According to the definition of color difference correlation, it can be
expressed as follows:
\begin{equation}
\label{eq8}
\phi ^{\left( k \right)}\left( {T_k \left( {\mbox{g}_c } \right)} \right) =
\frac{\sum {D\left( {T_k \left( {\mbox{g}_c } \right)} \right)} }{\sum {\bar
{D}\left( {T_k \left( {\mbox{g}_c } \right)} \right)} }
\end{equation}

It is a ratio of the color difference among perceptually uniform pixels to
that of all pixels. It can be seen from Eq.(\ref{eq5}) and Eq.(\ref{eq6}) that the more
the uniform neighbors are, the closer $D\left( {T_k \left( {\mbox{g}_c }
\right)} \right)$ is to $\bar {D}\left( {T_k \left( {\mbox{g}_c } \right)}
\right)$. Furthermore, when the number of uniform neighbors is fixed
($D\left( {T_k \left( {\mbox{g}_c } \right)} \right)$ is unchanged), the degree of
color difference of dissimilar pixels can be reflected. So color difference correlation $\phi ^{\left( k
\right)}$ characterizes the correlation information among pixels in local patches based on perceptually uniform color
difference.

\begin{center}
\begin{figure}[htbp]
\centerline{\includegraphics[width=3.2in]{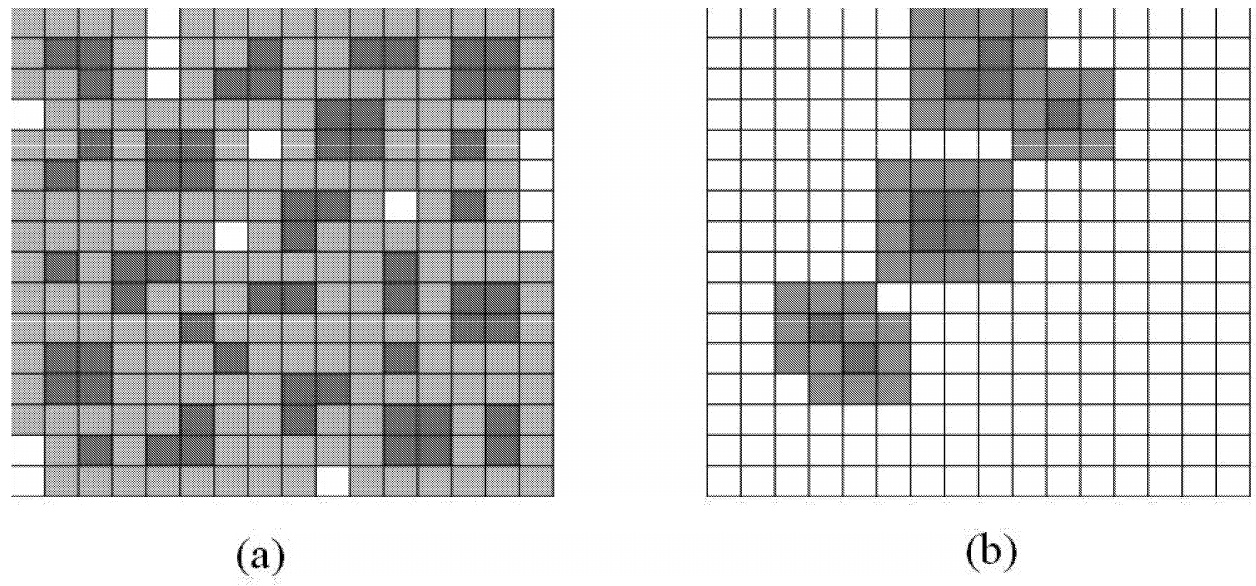}}
\caption{Two special image patterns}
\label{fig7}
\end{figure}
\end{center}

However, $\phi ^{\left( k \right)}$ overemphasizes the local feature in the patches, which
caused the loss of holistic characteristics. As shown in Fig. \ref{fig7}, there are
two different images with special contents where the red pixels represent certain property $T_k
$ (color or edge orientation). Provided that the color difference between
uniform pixels (both pixels are red) is set to $\bar {d}_1 $ while that
between non-uniform pixels (one is red and the other is green) is set
to $\bar {d}_2 $, color difference correlation feature in these images can be
calculated. In Fig. \ref{fig7}(a), the result is expressed as $\phi ^{\left( k
\right)}\left( {T_k } \right) = \frac{5\bar {d}_1 }{5\left( {\bar {d}_1 +
3\bar {d}_2 } \right)} = \frac{\bar {d}_1 }{\bar {d}_1 + 3\bar {d}_2 }$,
while that is $\phi ^{\left( k \right)}\left( {T_k } \right) = \frac{\bar
{d}_1 }{\bar {d}_1 + 3\bar {d}_2 }$ in Fig. \ref{fig7}(b). Apparently, these image are
significantly different even though they have identical results. Also it can
be seen that they have identical local patterns which is reason why they
show the same $\phi ^{\left( k \right)}$, but the occurrence frequencies of this local pattern are distinctive.

To solve the problem, we propose global color difference histogram to obtain
the global distribution probability of certain pattern. For the block where
the property of central pixel is $T_k \left( {\mbox{g}_c } \right)$, the global
color difference histogram can be described as:

\begin{equation}
\label{eq9}
\psi ^{\left( k \right)}\left( {T_k \left( {\mbox{g}_c } \right)} \right) =
\frac{\sum {\bar {D}\left( {T_k \left( {\mbox{g}_c } \right)} \right)} }{sum\left(
{\sum {\bar {D}\left( {T_k \left( {\mbox{g}_c } \right)} \right)} } \right)}
\end{equation}
where $\psi ^{\left( k \right)}$ denotes the probability that the actual color
distributions around the pixels which have property $T_k$ occur in holistic
regions. However, if only considering this feature, the local color
information is lost, which also limit the performance of the descriptor.

In order to extract image descriptor whose perceptually uniform color difference
characterizes the distribution in both local and holistic region, it is
necessary to fuse color difference correlation and global color difference
histogram. Even though the strategy that combines these feature with
appropriate weight may have their advantages simultaneously, it is possible
that these features cancel each other out in a way because they are highly
interrelated. In addition, the expansion of feature dimension may slow the
process of image retrieval. Therefore, color difference feature is defined
based on a method which has been presented to fuse the two features \cite{42}, and
can be expressed as follows:
\begin{equation}
\label{eq10}
L_c^{\left( k \right)} \left( {T_k \left( {\mbox{g}_c } \right)} \right) = \phi
^{\left( k \right)}\left( {T_k \left( {\mbox{g}_c } \right)} \right)\times \left(
{\psi ^{\left( k \right)}\left( {T_k \left( {\mbox{g}_c } \right)} \right) + 1}
\right)
\end{equation}

It can be seen from Eq.(\ref{eq10}) that the first term is color difference
correlation feature and the second term actually represent the global color
difference histogram. With the fusing method, $L_c^{\left( k \right)}$ has
the characteristics in both the two features to some degree. And feature
dimension does not increase, and maintain the length as $\phi ^{\left( k
\right)}$ and $\psi ^{\left( k \right)}$, which is important to improve the
speed of latter image retrieval. Besides that, color difference feature can
be rewritten as:
\begin{equation}
\label{eq11}
\begin{array}{l}
 L_c^{\left(k\right)} \left({T_k\left({\mbox{g}_c} \right)} \right) = \phi
^{\left( k \right)}\left( {T_k \left( {\mbox{g}_c } \right)} \right)\times \psi
^{\left( k \right)}\left( {T_k \left( {\mbox{g}_c } \right)} \right) + \phi
^{\left( k \right)}\left( {T_k \left( {\mbox{g}_c } \right)} \right) \\
 \;\;\;\;\;\;\;\;\;\;\;\;\;\;\;\;\; = \frac{\sum {D\left( {T_k \left( {\mbox{g}_c }
\right)} \right)} }{sum\left( {\sum {\bar {D}\left( {T_k \left( {\mbox{g}_c }
\right)} \right)} } \right)} + \frac{\sum {D\left( {T_k \left( {\mbox{g}_c }
\right)} \right)} }{\sum {\bar {D}\left( {T_k \left( {\mbox{g}_c } \right)}
\right)} } \\
 \end{array}
\end{equation}

It can be seen from Eq.(\ref{eq11}) that the first term is the percentage that
perceptually uniform color difference of pixels whose property are $T_k $ in
certain local patches accounts for color difference in the holistic regions,
and the second term is actually color difference correlation $\phi ^{\left(
k \right)}$, which represents the percentage that perceptually uniform color
difference of pixels in certain local patches accounts for all possible
color difference in these patches. Thus from this point of view,
$L_c^{\left( k \right)} $ characterizes the distribution of perceptually
uniform color difference in both local patches and holistic regions. Moreover, it
overcomes the problems in $\phi ^{\left( k \right)}$ and $\psi ^{\left( k
\right)}$. With this representation, two images as shown in Fig. \ref{fig7} can be
distinguished because of their completely different results in the first
term of Eq.(\ref{eq11}) even though they have the same $\phi ^{\left( k \right)}$.

\subsection{Texton frequency feature}

\begin{center}
\begin{figure}[htbp]
\centerline{\includegraphics[width=3.50in,height=1.40in]{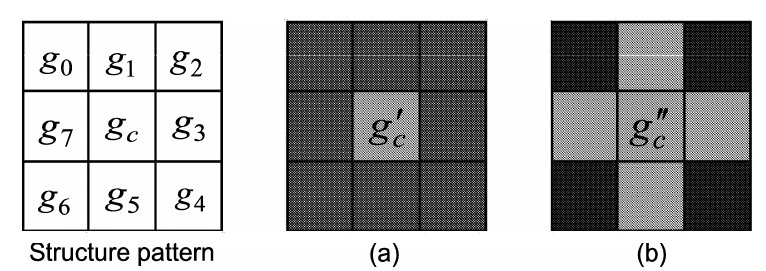}}
\caption{Two special structures which differ in color spatial correlation}
\label{fig8}
\end{figure}
\end{center}

Color difference feature aims at describing the contrast information among
pixels using the distribution of perceptually uniform color difference. As
discussed above, contrast is only one of the properties. Fig. \ref{fig8} contains two
special patterns which differ significantly in color spatial correlation
though they have the identical color difference for the center pixel. Thus
that only considering the color difference limits the discriminative
performance.

Spatial structure as the other property is orthogonal and complementary with
contrast, and can be expressed by the frequency distribution of uniform
pixels. Based on that, we propose the other feature called texton frequency
feature to describe the pattern types of pixels. To characterize the
frequency of pixels in local patches and holistic regions, texton frequency
histogram and texton frequency correlation are proposed in this section. And
then texton frequency feature is expressed through fusing them.

Histogram is one of the most common strategies to extract the global
occurrence probability of pixels. Provided that $\bar {N}\left( {T_k
\left( {\mbox{g}_c } \right)} \right)$ denotes the occurrence number of pixels
which have the property $T_k $, texton frequency histogram can be expressed as
follows:
\begin{equation}
\label{eq12}
\varphi ^{\left( k \right)}\left( {T_k \left( {\mbox{g}_c } \right)} \right) =
\frac{\bar {N}\left( {T_k \left( {\mbox{g}_c } \right)} \right)}{sum\left( {\bar
{N}\left( {T_k \left( {\mbox{g}_c } \right)} \right)} \right)}
\end{equation}
where $sum\left( \bar {N} \right)$ denotes the sum of number of pixels in
perceptual uniform image. $\varphi ^{\left( k \right)}$ describes the global
distribution of $T_k $, and concentrates on individual pixel regardless of
the relationship among them. Consequently, even though it computes
conveniently and demonstrates good performance in image analysis, histogram
statistical strategy still has limitations because of the loss of spatial
correlation.

To describe the spatial correlation among pixels, structure element
correlation (SEC) has been proposed in \cite{13}. Based on this idea, texton
frequency correlation can be written in following equation:

\begin{equation}
\label{eq13}
N\left( {T_k \left( {\mbox{g}_c } \right)} \right) = \sum {\sum\nolimits_i {\delta
_c \left( {\mbox{g}_c ,\mbox{g}_i } \right)} }
\end{equation}

\begin{equation}
\label{eq14}
\eta ^{\left( k \right)}\left( {T_k \left( {\mbox{g}_c } \right)} \right) =
\frac{\sum {\sum\nolimits_i {\delta _c \left( {\mbox{g}_c ,\mbox{g}_i } \right)} } }{8\bar
{N}\left( {T_k \left( {\mbox{g}_c } \right)} \right)}
\end{equation}
where $N\left( {T_k \left( {\mbox{g}_c } \right)} \right)$ counts the sum of uniform
neighbors $\mbox{g}_i $ which have the same property $T_k $ with the central
pixel $\mbox{g}_c $ in the perceptual uniform image. Thus $\eta ^{\left( k \right)}$
is a ratio of the actual number of uniform neighbors to all the possible
uniform neighbors. It characterizes the distribution of pixels in the local
patches.

That $\eta ^{\left( k \right)}$ emphasizes too much on the description in
local patches limits its performance to distinguish some images. Still take
Fig. \ref{fig7} for example, the red pixels mean that they have identical property $T_k
$. It can be seen that the occurrence frequency of red pixels are different
significantly. However, they have the same $\eta ^{\left( k \right)}$, which
is 0.25. The reason why texton frequency correlation does not distinguish
the two images is the occurrence frequency of $\eta ^{\left( k \right)}$
itself, which texton frequency histogram just emphasizes on. In order to
utilize the advantage of texton frequency histogram and texton frequency
correlation, texton frequency feature is proposed through the fusing method
as discussed in \cite{42}. Thus it can be expressed as follows:

\begin{equation}
\label{eq15}
L_f^{\left( k \right)} \left( {T_k \left( {\mbox{g}_c } \right)} \right) = \eta
^{\left( k \right)}\left( {T_k \left( {\mbox{g}_c } \right)} \right)\times \left(
{\varphi ^{\left( k \right)}\left( {T_k \left( {\mbox{g}_c } \right)} \right) + 1}
\right)
\end{equation}

The first term is the texton frequency correlation $\eta ^{\left( k
\right)}$ which describes the local distribution of $T_k $ and the second term can be seen as the texton frequency histogram $\varphi ^{\left( k
\right)}$ which characterizes the global distribution of $T_k $. Considering
utilizing the advantages both $\varphi ^{\left( k \right)}$ and $\eta ^{\left(
k \right)}$, $L_f^{\left( k \right)} $ describes the spatial structure of
pixels which have the same property $T_k $ in local patches and holistic
regions. And Eq.(\ref{eq12}) can be rewritten as follows:
\begin{equation}
\label{eq16}
\begin{array}{l}
 L_f^{\left( k \right)} \left( {T_k \left( {\mbox{g}_c } \right)} \right) = \varphi
^{\left( k \right)}\left( {T_k \left( {\mbox{g}_c } \right)} \right)\mbox{ +
}\varphi ^{\left( k \right)}\left( {T_k \left( {\mbox{g}_c } \right)} \right)\times
\eta ^{\left( k \right)}\left( {T_k \left( {\mbox{g}_c } \right)} \right) \\
 \;\;\;\;\;\;\;\;\;\;\;\;\;\;\;\;\;\;\mbox{ = }\frac{N\left( {T_k \left(
{\mbox{g}_c } \right)} \right)}{8\bar {N}\left( {T_k \left( {\mbox{g}_c } \right)}
\right)}\mbox{ + }\frac{N\left( {T_k \left( {\mbox{g}_c } \right)}
\right)}{8sum\left( {\bar {N}\left( {T_k \left( {\mbox{g}_c } \right)} \right)}
\right)} \\
 \end{array}
\end{equation}
where the first term is still texton frequency histogram $\varphi ^{\left( k
\right)}$, and the second term is a radio of the number of uniform pixels to
that of all possible uniform pixels, which represents the probability that
uniform pixels occur in global regions. With the second term, Fig. \ref{fig7}(a)(b) can be
distinguished even though they have the same $\varphi ^{\left( k \right)}$.

\subsection{Perceptual Uniform Descriptor (PUD)}

As described above, to extract two orthogonal and complementary properties,
color difference feature $L_c^{\left( k \right)} $ and texton frequency
feature $L_f^{\left( k \right)} $ are presented to characterize the contrast
and spatial structure among pixels in perceptually uniform regions,
respectively. Then combining these features, perceptual uniform descriptor
$H^k \left( {k = 1,2} \right)$ can obtain superior performance. Suppose that
contrast and spatial structure play equal part in discriminating images for
image retrieval, perceptual uniform descriptor can be expressed as:

\begin{equation}
\label{eq17}
H^k = \left[ {L_c^{\left( k \right)} \;\;L_f^{\left( k \right)} } \right]
\end{equation}
where $H^k$ is described based on perceptual uniform images $T_k$, and its corresponding dimension is $(128+12)=140$. $H^1$ and $H^2$ are color and edge orientation
uniform feature representation, respectively. Though both color and edge
orientation are important properties of images, they have different
performance in image datasets. To improve their performance, appropriate
weight is adopted to combine them. So the final feature representation can
be described as follows:

\begin{equation}
\label{eq18}
H = \left[ {\beta_1 \cdot H^1 \;\;\beta_2 \cdot H^2 } \right]
\end{equation}
where $\beta_1$ and $\beta_2$ denote the weight of $H_1$ and $H_2$,
respectively. And finally $H$ is a $140\times2=280$ dimensional feature vector.

\section{Manifold Ranking (MR)}

Manifold Ranking (MR) is a transductive ranking method which outperforms inductive ones in most
cases in CBIR. The notation and the ranking process of MR can be described in details as follows.

Given a set of features $H = \left\{ {H_1 ,H_2 , \cdots ,H_n } \right\} \in
{\mathds{R}}^{M\times n}$. Assuming the $q$-th image is the query. Let $d:H\times H
\to {\mathds{R}}$ is a map (metric) for each pair $H_i$ and $H_j$, where
$d\left( {H_i ,H_j } \right)$ is the distance between $H_i$ and $H_j$. We
denote $f=\left[{f_1 ,f_2 , \cdots ,f_n}\right]^T\in{\mathds{R}}^n$ as the
ranking results, where the ranking score $f_i$ corresponds to image $H_i$. The
initialized ranking score vector is defined by $y = \left[ {y_1 ,y_2 ,
\cdots ,y_n } \right]^T \in { \mathds{R}}^n$, where $y_q = 1$ if $H_q $ is the
query, and $y_q = 0$ otherwise.

An affinity matrix $W = \left\{ {W_{ij} } \right\} \in { \mathds{R}}^{n\times
n}$ which describes the distance of each pair of features on manifold is
defined as follows:
\begin{equation}
\label{eq19}
W_{ij} = \left\{ {\begin{array}{l}
 \exp \left( {{ - d^2\left( {H_i ,H_j } \right)} \mathord{\left/ {\vphantom
{{ - d^2\left( {H_i ,H_j } \right)} {2\sigma ^2}}} \right.
\kern-\nulldelimiterspace} {2\sigma ^2}} \right),H_i ,H_j \in N_K {\kern
1pt} \left( {H_l } \right){\kern 1pt} \\
 0,otherwise \\
 \end{array}} \right.
\end{equation}
where $N_K {\kern 1pt} \left( {H_q } \right)$ is the neighborhood of $H_q $,
and $K$ is the neighborhood parameter. Then, we define $S = D^{{ - 1}
\mathord{\left/ {\vphantom {{ - 1} 2}} \right. \kern-\nulldelimiterspace}
2}WD^{{ - 1} \mathord{\left/ {\vphantom {{ - 1} 2}} \right.
\kern-\nulldelimiterspace} 2}$ as symmetrically normalize matrix of $W$,
where $D$ is the diagonal matrix, $D_{ii} = \sum\nolimits_{j = 1}^n {W_{ij}
} $. In  Eq.(\ref{eq19}), $d\left( {H_i ,H_j } \right)$ may be L1-norm or L2-norm.

An iterate process is involved by the following equation.

\begin{equation}
\label{eq20}
f^{\left( {t + 1} \right)} = \alpha Sf^{\left( t \right)} + \left( {1 -
\alpha } \right)y
\end{equation}
where $\alpha $ is a parameter in $\left[ {0,1} \right)$. Since $\left( {I -
\alpha S} \right)$ is invertible, the direct method to compute the $f$ can
be expressed as follows:
\begin{equation}
\label{eq21}
f = \left( {I - \alpha S} \right)^{ - 1}y
\end{equation}
More details about MR are introduced in \cite{28,43}.

\section{The compatibility between PUD and manifold}
\begin{figure*}[htbp]
\centerline{
\includegraphics[width=7.5in]{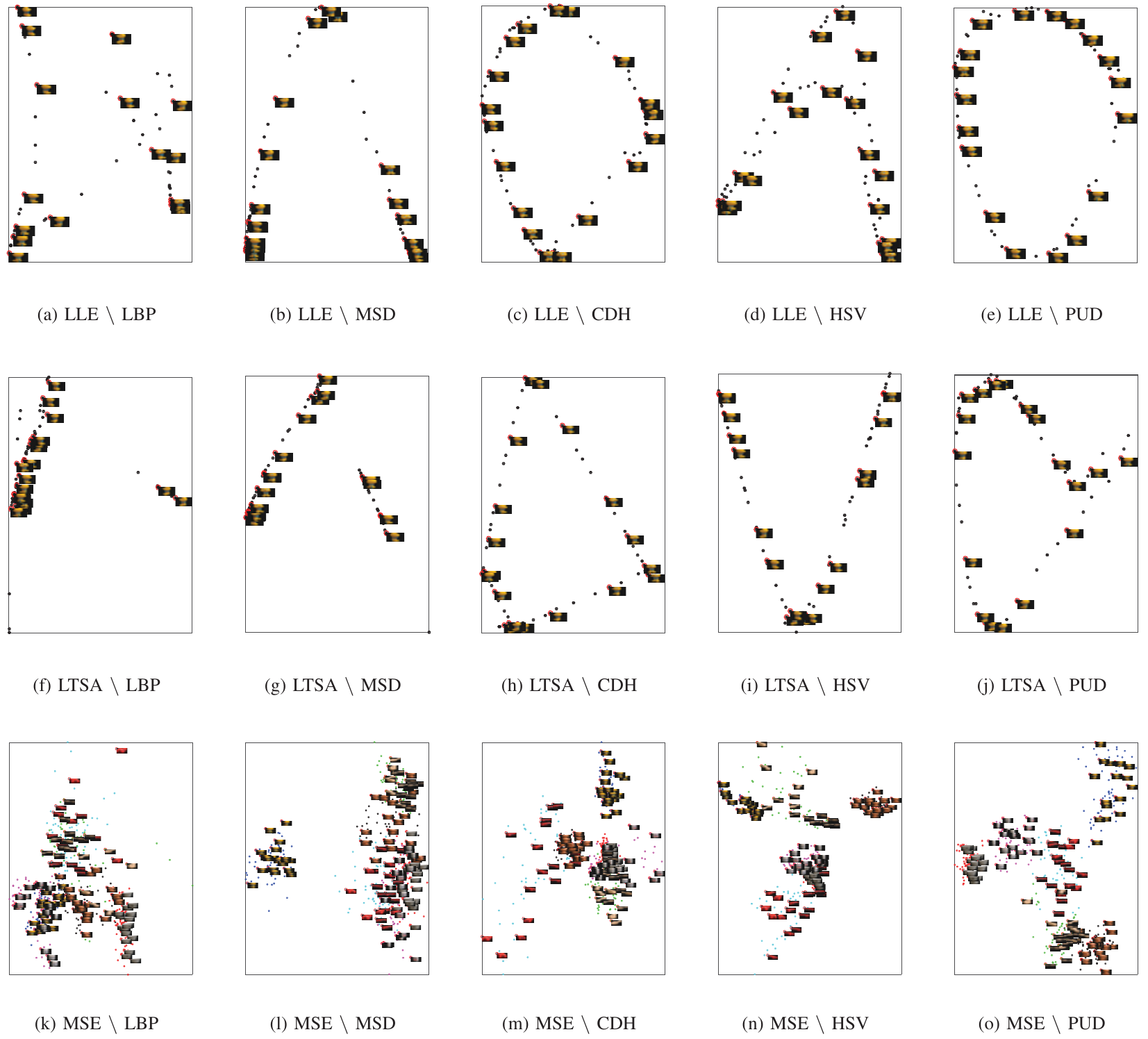}}
\caption{Manifold embedding in 2-D space of Coil100 dataset}
\label{fig1}
\end{figure*}

In coil100 dataset (see details in section \uppercase\expandafter{\romannumeral 7}), we employ
locally linear embedding (LLE) \cite{36}, local tangent space alignment (LTSA) \cite{54} and maximal similarity embedding (MSE) \cite{55} dimensionality reduction methods to give
visualizations of LBP, MSD, CDH, HSV histogram and PUD on 2-dimensional space, with neighborhood parameter $k=6$, as shown in Fig. \ref{fig1}. It can be seen from  Fig. \ref{fig1}(a)-(e) that a toy cat is captured by rotating from $0^{\circ}$ to $360^{\circ}$. LBP, MSD and HSV all fail to recover manifold structure while PUD recovers better manifold structure than the CDH one when utilizing LLE for manifold visualizations. And then image visualizations based on LTSA further prove this conclusion (shown in Fig. \ref{fig1}(f)-(j)). Furthermore, multi-class images are learnt based on MSE as shown in Fig. \ref{fig1}(k)-(o). It can be seen that PUD can clearly classify these five different kinds of images while all the other four descriptors confuse two or more types of images. Therefore, PUD is more related to manifold than other descriptor and suits for Manifold Ranking (MR).
\section{ Experimental Results}

Extensive experiments are conducted to test and illustrate the effectiveness
of our proposed scheme. In the experiments, we mainly compare our image
descriptor with local binary patterns (LBP) \cite{10}, micro-structure descriptor (MSD) \cite{13}, color difference histogram (CDH) \cite{14}and HSV color histogram \cite{53}. In ranking step, L1-norm, L2-norm, manifold ranking based on L1-norm (MR1) and based on L2-norm (MR2) are
involved. Some previous works related to our scheme are also considered, such
as Bag of Color, Bag of Words etc.

Besides, deep convolutional neural network (CNN) \cite{56} model is also analyzed and compared with our PUD to validate the effectiveness in image retrieval. It has been suggested that the features emerging in the middle of the fully-connected neural network layers hierarchy can serve as a high-level descriptor to extract the visual content of the image. This method has shown superior performance for large-scale image retrieval. Scatter balance metric learning (SBML) is introduced to measure the similarity among image features based on angle linear discriminant embedding (ALDE) \cite{55}. As a global dimensionality reduction method, ALDE aims to find transformation matrix $U$ to obtain low-dimensional projection $y=U^{T}x$ through scatter balance. With this idea, we can construct the measurement $dist(x_{1},x_{2})=\sqrt{x_{1}^{T}UU^{T}x_{2}}$ to analyze the performance of PUD with metric learning. In the experiments, we choose randomly 5\% images as training samples to construct $U$ of SBML in the datasets.
\subsection{Datasets}

In the experiments, Corel-1K, Corel-10K, Coil-100, UKbench and Cifar-10 datasets are utilized in
our CBIR scheme. Corel-1K is a small size dataset with only 1000
images. Others are large scale datasets. Corel-1K and Corel-10K are
involved to test category image retrieval. Coil-100, UKbench and Cifar-10 datasets are used to
evaluate instance image retrieval. The images of Coil-100 dataset rotate on
yoz space while the images rotate on xoy space in UKbench dataset. Scale changing is also involved in the two datasets. The Cifar-10 dataset is divided into five training batches and one test batch, and each batch has 10000 images in random order. The basic information of the five datasets are listed in Table \uppercase\expandafter{\romannumeral 1}.

\begin{table}[htbp]
\caption{ The attributes of experimental datasets}
\begin{center}
\begin{tabular}
{ccccc}
\hline
Dataset&
Image Size&
{\#} of Class&
{\#} of Each Class&
Total Image \\
\hline
Corel-1K&
384$\times$256&
10&
100&
1000
 \\
Corel-10K&
Vary&
100&
100&
10000
 \\
Coil-100&
128$\times$128&
100&
72&
7200
 \\
UKbench&
640$\times$480&
2550&
4&
10200
 \\
Cifar-10&
32$\times$32&
10&
6000&
60000
 \\
\hline
\end{tabular}
\end{center}
\label{tab1}
\end{table}

\begin{table*}[htbp]
\caption{  The precision and recall of different schemes with 20
returns in Corel-1K dataset(\%)}
\begin{center}
\begin{tabular}
{ccccccccccccc}
\hline
\raisebox{-1.50ex}[0cm][0cm]{Methods}&
\raisebox{-1.50ex}[0cm][0cm]{Performance}&
\multicolumn{11}{c}{Classes}  \\
\cline{3-13}
 &
 &
African&
Beach&
Building&
Bus&
Dinosaur&
Elephant&
Flower&
Horse&
Mountains&
Food&
Avg \\
\hline
\raisebox{-1.50ex}[0cm][0cm]{PUD-1-norm}&
Precision&
76.2&
42.05&
\textbf{83.00}&
91.25&
99.85&
66.90&
91.85&
92.95&
49.35&
88.95&
78.24 \\

 &
Recall&
15.24&
8.41&
16.60&
18.25&
19.97&
13.38&
18.37&
18.59&
9.87&
17.79&
15.65 \\

\raisebox{-1.50ex}[0cm][0cm]{PUD-2-norm}&
Precision&
75.85&
50.55&
72.85&
94.10&
99.15&
64.65&
88.40&
89.90&
47.95&
83.25&
76.67 \\
 &
Recall&
15.17&
10.11&
14.57&
18.82&
19.83&
12.93&
17.68&
17.98&
9.59&
16.65&
15.33 \\
\raisebox{-1.50ex}[0cm][0cm]{PUD-MR1}&
Precision&
83.95&
43.70&
78.65&
\textbf{94.30}&
99.55&
73.80&
\textbf{98.40}&
96.70&
56.50&
\textbf{92.10}&
\textbf{81.77} \\
 &
Recall&
16.79&
8.74&
15.73&
18.86&
19.91&
14.76&
19.68&
19.34&
11.3&
18.42&
16.35 \\
\raisebox{-1.50ex}[0cm][0cm]{PUD-MR2}&
Precision&
80.70&
53.95&
71.85&
91.75&
98.05&
68.65&
96.35&
92.30&
55.00&
89.65&
79.83 \\
 &
Recall&
16.14&
10.79&
14.37&
18.35&
19.61&
13.73&
19.27&
18.46&
11.00&
17.93&
15.97 \\
\raisebox{-1.50ex}[0cm][0cm]{Guo et al.\cite{44}}&
Precision&
\textbf{84.70}&
45.40&
67.80&
85.30&
99.30&
71.10&
93.30&
95.80&
49.80&
80.80&
77.30 \\
 &
Recall&
16.94&
9.08&
13.56&
17.06&
19.86&
14.22&
18.66&
19.16&
9.96&
16.16&
15.46 \\
\raisebox{-1.50ex}[0cm][0cm]{Walia et al.\cite{45}}&
Precision&
51.00&
\textbf{90.00}&
58.00&
78.00&
78.00&
\textbf{100.00}&
84.00&
\textbf{100.0}&
\textbf{84.00}&
38.00&
78.30 \\
 &
Recall&
10.20&
18.00&
11.60&
15.60&
15.60&
20.00&
16.80&
20.00&
16.80&
7.60&
15.66 \\
\raisebox{-1.50ex}[0cm][0cm]{GMM\cite{46}}&
Precision&
72.50&
65.20&
70.60&
89.20&
\textbf{100.00}&
70.50&
94.80&
91.80&
72.25&
78.80&
80.57 \\
 &
Recall&
14.50&
13.04&
14.12&
17.84&
20.00&
14.10&
18.96&
18.36&
14.45&
15.76&
16.11 \\
\hline
\end{tabular}
\end{center}
\label{tab4}
\end{table*}

\begin{table*}[htbp]
\caption{ The precision comparison of different schemes with 20 returns in
Corel-1K dataset(\%)}
\begin{center}
\begin{tabular}
{ccccccccccccc}
\hline
\raisebox{-1.50ex}[0cm][0cm]{Methods}&
\multicolumn{11}{c}{Classes}  \\
\cline{2-12}
 &
African&
Beach&
Building&
Bus&
Dinosaur&
Elephant&
Flower&
Horse&
Mountains&
Food&
Avg \\
\hline
PUD-MR1&
83.95&
43.70&
\textbf{78.65}&
\textbf{94.30}&
99.55&
\textbf{73.80}&
\textbf{98.40}&
\textbf{96.70}&
56.50&
\textbf{92.10}&
\textbf{81.77 }\\
Yu\cite{47}&
\textbf{84.90}&
35.60&
61.60&
81.80&
\textbf{100.00}&
59.10&
93.10&
92.80&
40.40&
68.20&
71.70 \\
Lin\cite{48}&
68.30&
54.00&
56.20&
88.80&
99.30&
65.80&
89.10&
80.30&
52.20&
73.30&
72.70 \\
SBML&
74.00&
\textbf{60.05}&	
75.65&	
93.85&	
99.00&	
65.85&	
94.05&	
86.10&	
\textbf{59.90}&	
86.45&
79.49\\
\hline
\end{tabular}
\end{center}
\label{tab5}
\end{table*}

\begin{table*}[htbp]
\caption{ The precision and recall of different schemes with 12 returns in Corel-10K dataset(\%)}
\begin{center}
\begin{tabular}
{cccccccc}
\hline
Performance $\backslash$ Method&
SSH \par \cite{49}&
Ri-HOG \par \cite{50}&
HOG \par \cite{12}&
CNN&
LBP-MR1&
MSD-MR1&
PUD-MR1 \\
\hline
Precision& 54.88&
52.13 &
33.29&
49.63&
35.84&
49.65&
\textbf{58.46} \\
Recall & 6.58&
6.25&
3.94&
5.96&
4.30&
5.96&
7.02 \\
\hline
\end{tabular}
\end{center}
\label{tab7}
\end{table*}

\begin{figure*}[htbp]
\centerline{\includegraphics[width=7.5in]{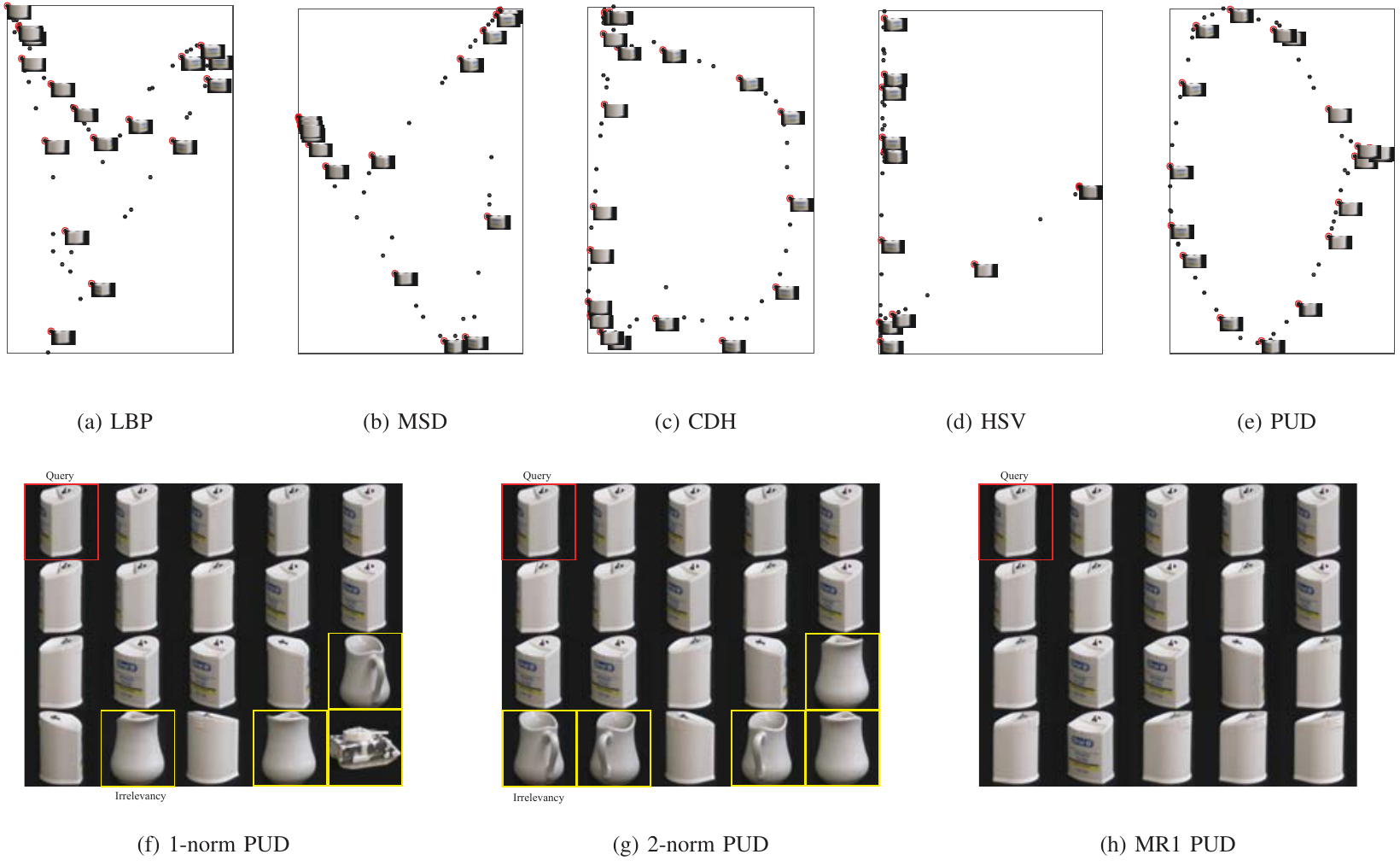}}
\caption{The manifold embedding and retrieval results of Coil-100 ($q\mbox{ =
591})$ dataset}
\label{fig11}
\end{figure*}

\begin{table*}[htbp]
\caption{ The N-S score of different schemes in UKbench dataset}
\begin{center}
\begin{tabular}
{cccccccccc}
\hline
Feature methods &
Guo etal. \par \cite{44}&
BOW-SIFT\cite{51}&
HE-SIFT \par \cite{51}&
BOC \par \cite{52}&
LBOC \par \cite{52}&
BOW-c-MI-Burst \par \cite{53}&
HSV&
CNN&
PUD-MR1 \\
\hline
N-S score &
3.42&
2.88&
2.58&
3.34&
3.50&
3.52&
3.40&
3.51&
\textbf{3.58} \\
\hline
\end{tabular}
\end{center}
\label{tab10}
\end{table*}

\begin{figure*}[htbp]
\centerline{\includegraphics[width=7.5in]{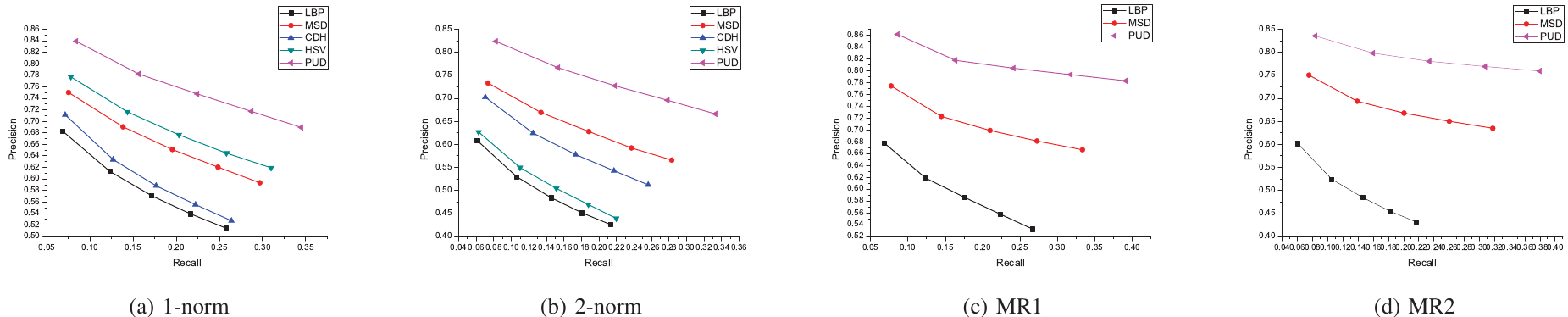}}
\caption{The performance comparation of image descriptors and ranking methods in Corel-1K dataset}
\label{fig12}
\end{figure*}

\begin{figure*}[htbp]
\centerline{\includegraphics[width=7.5in]{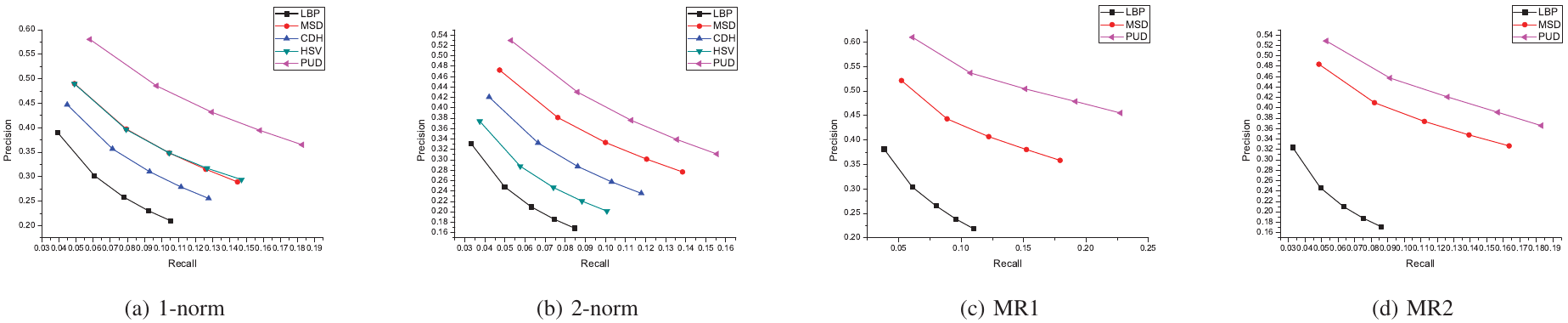}}
\caption{The performance comparation of image descriptors and ranking methods in Corel-10K dataset}
\label{fig13}
\end{figure*}

\begin{figure*}[htbp]
\centerline{\includegraphics[width=7.5in]{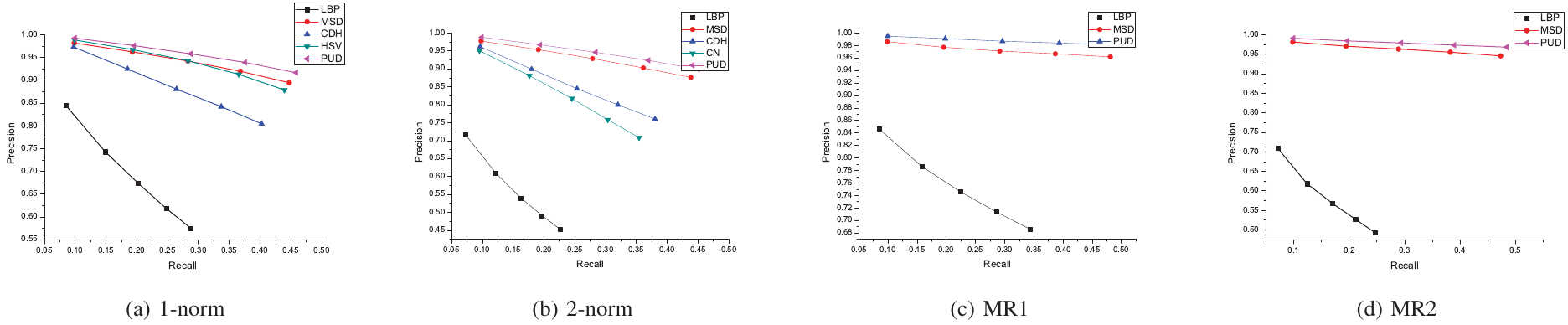}}
\caption{The performance comparation of image descriptors and ranking methods in Coil-100 dataset}
\label{fig14}
\end{figure*}

\begin{table*}[htbp]
\caption{ The precision of different image descriptors and ranking methods
with 400 returns in Cifar-10 dataset (\%)}
\begin{center}
\begin{tabular}
{ccccccccc}
\hline
Methods &
LBP 2-norm&
LBP MR2&
HSV 2-norm&
HSV MR2&
PUD 2-norm&
PUD MR2 &
CNN &
SBML\\
\hline
Precision&
20.58	&
14.14	&
19.29	&
19.41	&
27.51	&
\textbf{32.65}	&
29.96&
27.55\\
\hline
\end{tabular}
\end{center}
\label{tab11}
\end{table*}

\subsection{Experimental analysis}

In order to ensure the fairness of the experiments, we use the basic parameters (see Table  \uppercase\expandafter{\romannumeral 2}, where $\beta_1, \beta_2\in[0,2]$) and compare PUD with the other holistic features. And their feature dimensions are listed in Table \uppercase\expandafter{\romannumeral 3}. According to the length of feature vector, the dimension of PUD is reduced to 200 when utilizing SBML to measure the similarity and rank in image retrieval. And the dimension of LBP, MSD, CDH, HSV are reduced to 200, 50, 50 and 200 respectively. CNN is used to validate the performance of PUD in large-scale image datasets. The four basic holistic image features are also involved by manifold ranking, and the retrieval results illustrate the effectiveness of our proposed method. In our experiments, all the images in each dataset are alternated as query image. The definition of precision and recall used to evaluate the Corel-1K, Corel-10K, Coil-100 and Cifar-10 datasets are the same as in reference \cite{49}. N-S score (the best is 4) is the quantitative evaluation for UKbench dataset in our image retrieval task.
\begin{table}[htbp]
\caption{The parameter settings in the experimental datasets}
\begin{center}
\begin{tabular}
{cccccc}
\hline
Dataset $\backslash$ Parameter &
$\beta_1$&
$\beta_2$&
$K$&
$\alpha $&
$\sigma $ \\
\hline
Corel-1K&
0.1&
0.75&
8&
0.95&
2 \\
Corel-10k&
1&
1.65&
8&
0.95&
2 \\
Coil-100&
1&
1.65&
8&
0.95&
2 \\
UKbench&
1&
1&
8&
0.5&
2 \\
Cifar-10&
0.1&
0.75&
8&
0.95&
2 \\
\hline
\end{tabular}
\end{center}
\label{tab2}
\end{table}

\begin{table}[htbp]
\caption{ The dimensions of different image descriptors}
\begin{center}
\begin{tabular}
{ccccccc}
\hline
Method&
LBP&
MSD&
CDH&
HSV&
CNN&
PUD \\
\hline
Dim&
256&
72&
90&
1000&
4096&
280 \\
\hline
\end{tabular}
\end{center}
\label{tab3}
\end{table}

In Corel-1K dataset with 20 returns, our scheme (PUD-MR1) achieves better
average precision than others as shown in Table \uppercase\expandafter{\romannumeral 4}, \uppercase\expandafter{\romannumeral 5} and Fig. \ref{fig12}, where PUD-MR1 outperforms other methods in Bus, Flower, and Food classes as reported in Table \uppercase\expandafter{\romannumeral 4}. The precision of PUD with L1-norm ranking can reach 83.00{\%} which is
better than that of PUD-MR1 (78.65{\%}) in building (deduced by 4.35{\%}). This result
illustrates that the distribution of some natural images may not on the manifold and thus the
ranking score will not be corrected by perception of manifold. Table \uppercase\expandafter{\romannumeral 5} also
illustrates that our method is more effective than other state-of-the-art methods in Corel-1K dataset. Though the average precision of PUD-SBML model reaches 79.49\%, which is obviously higher than most of the methods, PUD-MR1 still shows better performance than PUD-SBML (+2.28\%) for image retrieval. CNN model is not suitable for small-scale image dataset because of the under-fitting in neural network, and thus is not compared and analyzed with our scheme in this dataset. These results demonstrate the rationality and effectiveness of the compatibility between PUD and manifold ranking in CBIR.

In Table \uppercase\expandafter{\romannumeral 6}, we compare our scheme with LBP, MSD, CDH and HSV in Corel-10K dataset.
Corel-10K dataset is an extend version of Corel-1K. As shown in Table \uppercase\expandafter{\romannumeral 6}, \uppercase\expandafter{\romannumeral 7} and Fig. \ref{fig13}, the average precision of LBP with L1-norm is slightly higher than that with the MR1 one (+0.66{\%}).
The L2-norm-related methods also show the similar results. This is mainly
because LBP is texture-based descriptor, which causes the image features
are not distributed on manifolds. In contrast, MSD and PUD relate to visual
uniformity of human. Based on our analysis, the image representations of MSD and PUD
can proximately embed on image manifold, which leads to better performance with
MR than that with norm-related ranking. Furthermore, MSD considers less visual
uniformity than PUD, this is also shown in Table \uppercase\expandafter{\romannumeral 9}. CDH and HSV cannot
perform on MR, which may mainly because the two descriptors are not
distributed on manifold. CDH and HSV may also encounter singular matrix of
graph on MR as a computational problem. This conclusion is also proved by
the analysis of Fig. \ref{fig1} and Fig. \ref{fig11}. In Corel-10K dataset, we also compare our results with
Ri-HOG and HOG. Table \uppercase\expandafter{\romannumeral 7} shows that our method outperforms Ri-HOG and HOG because HOG-based descriptors contain no color and texture information of images. Besides, the average precisions of SSH and CNN are also 3.58\% and 8.83\% lower than our PUD-MR1 in this dataset, respectively.
\begin{table}[htbp]
\caption{ The precision and recall of different image descriptors and ranking
methods with 12 returns in Corel-10K dataset (\%)}
\begin{tabular}
{cccccc}
\hline
Methods&
Performance Type&
1-norm&
2-norm&
MR1&
MR2 \\
\hline
LBP&
Precision&
36.50&
30.64&
35.84&
30.22 \\
&
Recall&
4.38&
3.68&
4.3&
3.63 \\
MSD&
Precision&
46.41&
44.77&
49.65&
46.01 \\
&
Recall&
5.57&
5.37&
5.96&
5.52 \\
CDH&
Precision&
42.08&
39.6&
-&-
 \\
&
Recall&
5.05&
4.76&
-&-
 \\
HSV&
Precision&
46.33&
34.87&
-&-
 \\
&
Recall&
5.56&
4.18&
-&-
 \\
PUD&
Precision&
55.51&
50.24&
\textbf{58.46}&
50.63 \\
&
Recall&
6.66&
6.03&
7.02&
6.08 \\
\hline
\end{tabular}
\label{tab6}
\end{table}

The experiments in Coil-100 dataset can give an intuitive interpretation why the combination between PUD
and MR realizes perceptual uniformity. Table \uppercase\expandafter{\romannumeral 8} and Fig. \ref{fig14} show the results in
Coil-100 dataset. Compared with other descriptors, the experimental results of
LBP show lower precision because Coil-100 is a color image dataset. LBP only
involves image texture while others involve image color in descriptor.
From Fig. \ref{fig1} and Fig. \ref{fig11} (the 9-th class), we can see PUD is embedded better than
other descriptors on manifold by LLE method. Furthermore, we explain why
MR-based methods are better than norm-based ranking ones. In Coil-100
dataset, let $q = 591$ as a query. The scores of relevant samples in
norm-based ranking methods are independent while that of MR-based methods are
propagated by graph matrix $S$. Therefore, some irrelevant samples may
``similar'' with the query as a correct result, which may not occur with MR.
Fig. \ref{fig1} and Fig. \ref{fig11} also demonstrate that PUD gets better visual uniformity than LBP, MSD,
CDH and HSV. In Fig. \ref{fig1}(d) and Fig. \ref{fig11}(d), HSV involves no visual uniformity, which is proved by
the embedding results of LLE. Besides, SBML-based methods are compared and analyzed in this part. It can be seen from Table \uppercase\expandafter{\romannumeral 8} that SBML-based methods are worse than norm-based or MR-based ones.
\begin{table}[htbp]
\caption{ The precision of different image descriptors and ranking methods
with 20 returns in Coil-100 dataset (\%)}
\begin{center}
\begin{tabular}
{cccccc}
\hline
Methods &
1-norm&
2-norm&
MR1&
MR2 &
SBML\\
\hline
LBP&
74.30&
60.95&
78.69&
61.87&
58.35 \\
MSD&
96.25&
95.38&
97.72&
97.09&
95.20 \\
CDH&
92.48&
89.92&
-&
- &
90.96\\
HSV&
96.73&
88.10&
-&
- &
89.07\\
PUD&
97.63&
96.74&
\textbf{99.11}&
98.42&
93.69 \\
\hline
\end{tabular}
\end{center}
\label{tab8}
\end{table}

UKbench dataset is not suitable for MR because only four samples in each class.
However, our scheme gets better performance than LBP, MSD, CDH and HSV (shown in Table \uppercase\expandafter{\romannumeral 9}). The change in N-S score of PUD +0.22 is a competitive performance on manifold ranking (LBP: -0.24, MSD: +0.01). As shown in Table \uppercase\expandafter{\romannumeral 10}, the N-S score 3.58 of PUD-MR1 also outperforms SIFT-based and BOC-based local descriptors.
\begin{table}[htbp]
\caption{ The N-S score of different image descriptors and ranking methods in
UKbench dataset}
\begin{center}
\begin{tabular}
{ccccc}
\hline
Methods&
1-norm&
2-norm&
MR1&
MR2 \\
\hline
LBP&
1.84&
1.61&
1.60&
1.42 \\
MSD&
3.23&
3.11&
3.24&
2.94 \\
CDH&
2.49&
2.35&
-&
- \\
HSV&
3.40&
3.20&
-&
- \\
PUD&
3.36&
3.30&
\textbf{3.58}&
3.45 \\
\hline
\end{tabular}
\end{center}
\label{tab9}
\end{table}

\begin{center}
\begin{figure}[htbp]
\centerline{\includegraphics[width=3.5in]{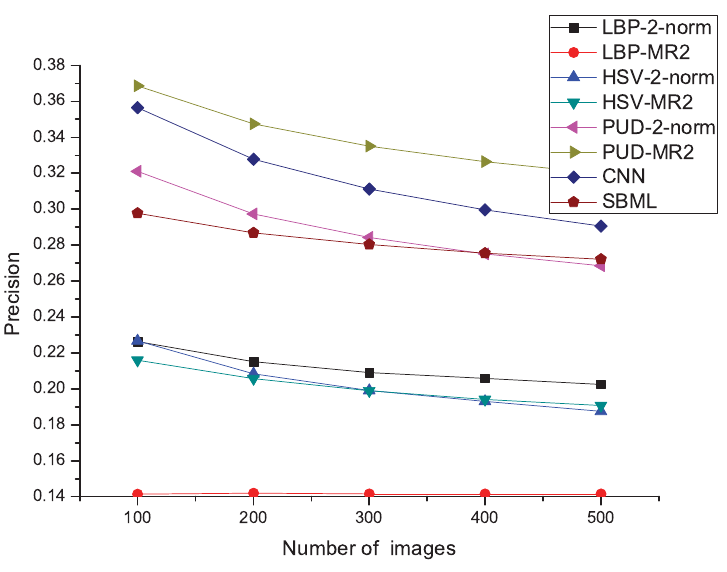}}
\caption{The performance comparation of image descriptors and ranking methods in Cifar-10 dataset}
\label{fig15}
\end{figure}
\end{center}

As one of the large scale widely datasets, Cifar-10 dataset also utilized to test the performance of our method compared in with LBP, HSV, CNN and PUD-SBML. As shown in Fig. \ref{fig15} and Table \uppercase\expandafter{\romannumeral 11}, PUD-MR2 outperforms other methods with the increase of returning images though CNN and PUD-SBML also present superior performance in the large-scale dataset. When returning 400 images in this dataset, the average precision of PUD-MR2 can reach 32.65\%, which is 2.99\% and 5.1\% higher than CNN and PUD-SBML, respectively. It also demonstrates that the combination between PUD and MR has better retrieved results than norm-based method. Therefore, the results in the five datsets illustrate that the compatibility between image representation and ranking based on visual and perceptual uniformity plays an important role in image retrieval.

\section{Conclusion}
In this paper, an effective holistic image feature extraction method is
proposed based on Gestalt psychology, namely Perceptual Uniform Descriptor. By
manifold learning method and visualization, we proved that our descriptor is
more suitable to use manifold ranking than other descriptors mentioned in
this paper. Furthermore, the experimental results show that the combination
between PUD and manifold ranking is effective for image retrieval in most cases.
However, in few cases, the L1-norm ranking obtains a higher accuracy than
the manifold one. This phenomenon shows that the images in the dataset are not
distributed on a manifold, and the effectiveness of manifold ranking is not
satisfactory. Finally, we point out that the compatibility between image
descriptors and ranking method is a very significant problem. Descriptors
and ranking are equally important, and should be considered as a framework
in image retrieval task.

In our future work, the manifold ranking method for multi-graph fusion and the construction of a robust graph will be considered. In terms of image descriptor,
the combination between global and local representations still should be
further explored. Besides, cognitive psychology and human visual system
need to be researched and applied to detect the objects or scenes.



\ifCLASSOPTIONcaptionsoff
  \newpage
\fi

\end{document}